 \documentclass[final,5p,times,twocolumn,authoryear]{elsarticle}

\usepackage{amssymb}
\usepackage{lipsum}
\usepackage{amsmath}
\usepackage{amsfonts}
\usepackage{algorithmic}
\usepackage{graphicx}
\usepackage{textcomp}
\usepackage{xcolor}
\usepackage{algorithm}
\usepackage{amsmath}
\usepackage{multirow}
\usepackage{tabularx}
\usepackage{mathtools, nccmath}
\usepackage{booktabs}
\usepackage{subfig}

\begin{document}

\begin{frontmatter}



\title{Overcoming Growth-Induced Forgetting with Sparsity in Task-Agnostic Continual Learning}





\author[1]{Yuqing Zhao}
\author[1]{Jiannong Cao}
\author[3]{Divya Saxena}
\author[1]{Xiaoyun Liu}
\author[1]{Changlin Song}
\author[6]{Bo Yuan}
\author[7]{Julie McCann}

\affiliation[1]{Hong Kong Polytechnic University}
\affiliation[3]{Indian Institute of Technology}
\affiliation[6]{China Mobile Hong Kong}
\affiliation[7]{Imperial College London}

\begin{abstract}
In continual learning (CL), model growth enhances adaptability to new data. However, when model growth is applied improperly—especially in task-agnostic CL where the entire grown model is used for inference—it can lead to severe degradation of learned knowledge, a problem we term growth-induced forgetting. Most existing methods that adopt model growth to improve adaptability often overlook the forgetting issue, resulting in compromised knowledge retention, making them unsuitable for task-agnostic settings. To promote both adaptability and knowledge retention with model growth, we identify the key: gradient and parameter sparsity. Introducing SparseGrow, which increases gradient sparsity through layer expansion and gradient gating to enable focused updates on parameters while preserving critical parameters, thus inhibiting forgetting. Moreover, it promotes parameter sparsity with sparse initialization and training, aiming at better control of model plasticity, improving adaptability over new data. Extensive experiments across diverse datasets, task-agnostic settings, and a large number of tasks demonstrate the necessity of controlled layer expansion and validate the effectiveness of SparseGrow in achieving high adaptability while minimizing forgetting in continual learning. By enabling model growth with sparsified gradient and parameters, SparseGrow paves the way for building scalable lifelong learning systems capable of continual adaptation with better knowledge retention.
\end{abstract}



\begin{keyword}
Task-Agnostic Continual Learning \sep Sparsity \sep Growth-Induced Forgetting \sep Model Growth\sep Adaptability


\end{keyword}

\end{frontmatter}




\section{Introduction}
\label{introduction}









\begin{figure}[!ht]
  \begin{center}
{\includegraphics[width=\linewidth]{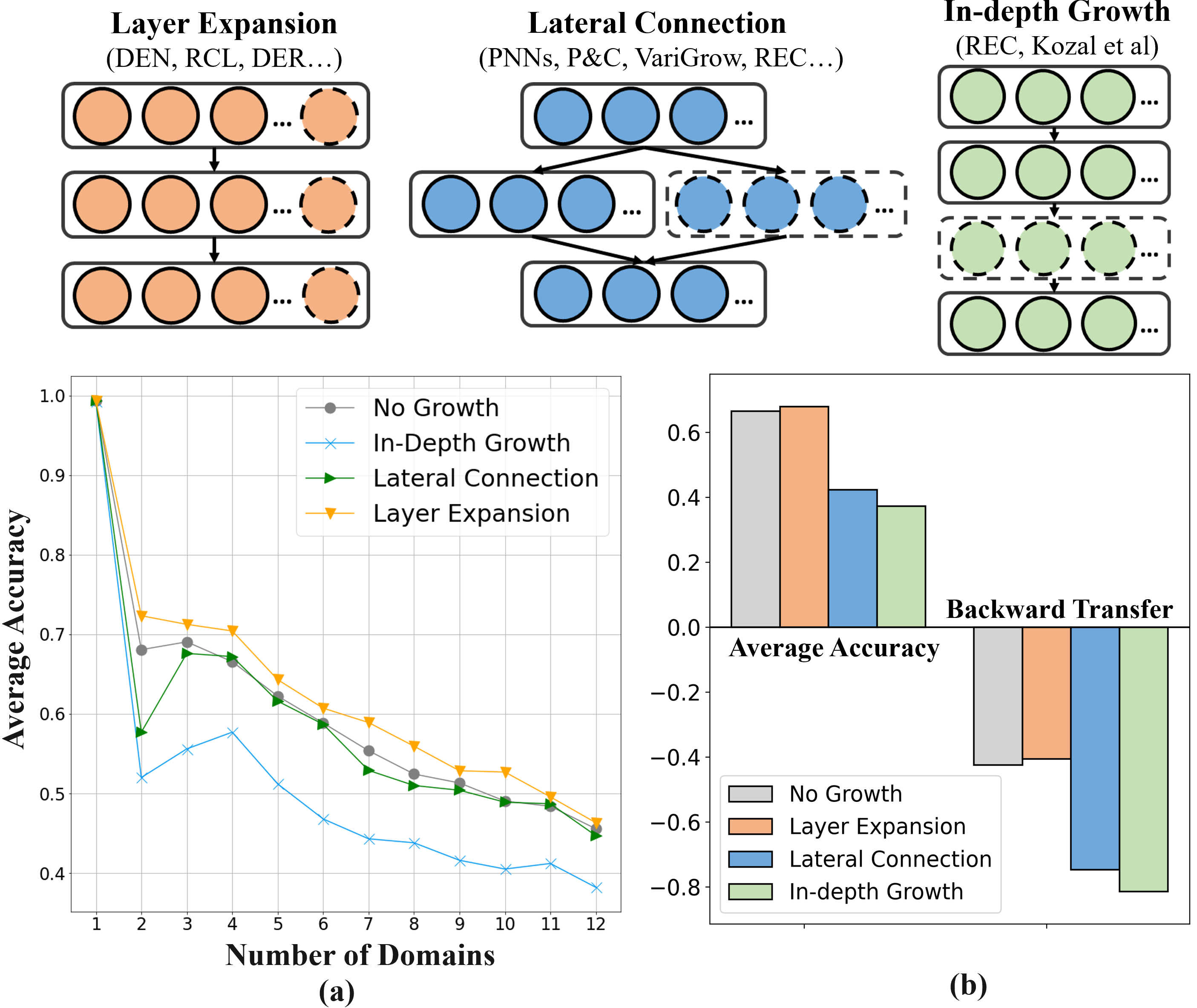}}
  \\
  \caption{Comparison of three model growth strategies applied to ResNet in task-agnostic continual learning (CL) across sequential domains, along with a baseline without model growth (No Growth).
The strategies include:
Layer Expansion: Widening existing layers (e.g., DEN, RCL, DER) to \textbf{increase gradient sparsity} and resist forgetting.
Lateral Connections: Adding new lateral layers (e.g., PNNs, VariGrow, P\&C, REC) for model adaptability.
In-Depth Growth: Increasing model depth via added hidden layers (e.g., REC, Kozal et al.) for enhanced adaptability.
(a) shows the change in average accuracy across domains; (b) reports both average accuracy and backward transfer, a measure of forgetting.
These results provide initial evidence for our study of growth-induced forgetting in task-agnostic CL. While Layer Expansion demonstrates superior accuracy and minimal forgetting by increasing gradient sparsity, other growth strategies may lead to higher degradation of past knowledge. This underscores the importance of controlled model growth in scenarios where the entire model is used for inference across evolving tasks. (Best viewed in color.)}
     \label{figure1}
  \end{center}
\end{figure}
\setlength{\textfloatsep}{10pt plus 1.0pt minus 2.0pt}

In the field of continual learning (CL), model growth becomes essential for better adaptation to new data and experiences, thereby improving scalability for managing incremental knowledge \cite{hanmo2024effective,chu2023continual}. In real-life scenarios with evolving needs—such as self-evolving agents, autonomous vehicles, and language models—model growth is crucial for effectively handling information updates, increasing task complexity, and incorporating new features or modalities. As experiences accumulate and task complexity rises, a fixed-capacity model can quickly become a bottleneck, struggling to integrate new information while retaining the knowledge it has previously acquired. To address the above challenge, the model needs to intelligently expand its parameter capacity. This helps ensure high adaptability and sustainable scalability, forming the foundation for truly lifelong learning systems.

However, improper model growth often leads to a degradation of performance on previously learned tasks, an issue we identify as growth-induced forgetting, especially in task-agnostic continual learning settings such as domain and class incremental learning settings \cite{zhao2023adaptcl,shim2021online,hu2021continual}. In these scenarios, the task label is unavailable, and the entire model, including expanded parameters, are used for inference. 
\textit{Unlike traditional catastrophic forgetting, which occurs due to changes in data distribution, \textbf{growth-induced forgetting} is a distinct phenomenon. It arises not from new data but from increasing the model’s parameter capacity—for example, by adding new neurons or layers to a previously trained model. This growth can disrupt the model’s ability to retain or effectively use the knowledge learned before the growth.}
A similar phenomenon is observed in neuroscience, where increasing neurogenesis in the brain after memory formation can promote forgetting, referred to as neurogenesis-based forgetting \cite{davis2017biology}.

Some existing works in continual learning \cite{wen2020autogrow,yang2021grown,kozal2023increasing} adopt different model growth strategies and randomly initialize them to enhance adaptability. However, they fail to recognize the presence of growth-induced forgetting resulting from improper model growth.
These methods are mostly task-specific, requiring manual task identification during inference to avoid forgetting, and not applicable for domain or class incremental scenarios; the rest are task-agnostic methods adopting improper model growth and initialization strategies that may further lead to growth-induced forgetting.
Nevertheless, the lack of explicit acknowledgment of the growth-induced forgetting issue poses challenges in identifying suitable model growth approaches. This not only limits comprehensive control of forgetting but also hinders full utilization of model growth.


To our knowledge, this study is the first to identify and systematically investigate the phenomenon of growth-induced forgetting in continual learning. Our work provides a comprehensive analysis aimed at developing effective strategies for model growth that expand the model's capacity while minimizing the risk of forgetting previously learned knowledge.
Our study finds that layer expansion, which widens layers without impacting model functionality, can increase the sparsity of the model gradient in subsequent learning \cite{mirzadeh2022wide}. This feature distinguishes it from alternative model growth strategies, such as lateral connections and in-depth growth. 

The impacts of these growth methods on growth-induced forgetting differ: Layer expansion expands parameters in width, which are computed collectively during inference;
Lateral connections here refer to the addition of lateral layers, with the values from the lateral layers added to the main pathway; In-depth growth involves adding new hidden layers to a neural network, introducing new computations for preceding and succeeding layers' parameters.
 To promote both adaptability and knowledge retention with model growth, we have identified two critical elements: gradient and parameter sparsity.
Drawing inspiration from neurogenesis \cite{davis2017biology,deng2010new}, we propose a novel sparse model expansion growing (SparseGrow) approach.
SparseGrow utilizes layer expansion and gradient gating to improve gradient sparsity. With expanded layers, only a small portion of the model's parameters need to be updated when training for a new task. Meanwhile, gradient gating protects important parameters by using gradient masks. This sparsified gradient allows for targeted updates on the parameters while maintaining critical ones, ensuring network stability and preventing forgetting caused by growth.
Furthermore, SparseGrow enhances parameter sparsity through both sparsity in training and initialization. Sparse training helps maintain a compact network without compromising accuracy, allowing the network to retain parameters that can adapt to new tasks. By initializing expanded layers with partially zero-valued parameters that are tailored to the dataset distribution, we aim to carefully control the model's plasticity. This approach enhances adaptability to new data and helps reduce forgetting that can occur due to model growth. In contrast, zero initialization can hinder updates, while random initialization may introduce unwanted interference.

To validate our findings, this study conducted extensive experiments across different task-agnostic settings using domain and class incremental datasets with varying numbers of tasks. The results highlight the importance of gradient and parameter sparsity during layer expansion. Our approach effectively addresses the issue of growth-induced forgetting while demonstrating adaptability and knowledge retention for incremental tasks.
The main contributions of this paper are threefold:

\begin{itemize}
    \item Our study reveals growth-induced forgetting, a novel variant of catastrophic forgetting observed when a model's capacity expands. This discovery fills a gap in existing research and contributes to the ongoing research on catastrophic forgetting.
    \item This work presents an in-depth study comparing various model growth strategies, revealing the potential of layer expansion to retain model knowledge through increased gradient sparsity.
    \item This work highlights the critical roles of gradient and parameter sparsity in continual learning. We propose SparseGrow, a novel method that effectively mitigates growth-induced forgetting via layer expansion, gradient gating, and sparse initialization/training during model growth. Our study shows that SparseGrow effectively maintains adaptability and knowledge retention in task-agnostic continual learning scenarios, supported by experimental validation in both domain and class-incremental settings.
\end{itemize}






\begin{figure*}[!ht]
  \begin{center}
{\includegraphics[width=\linewidth]{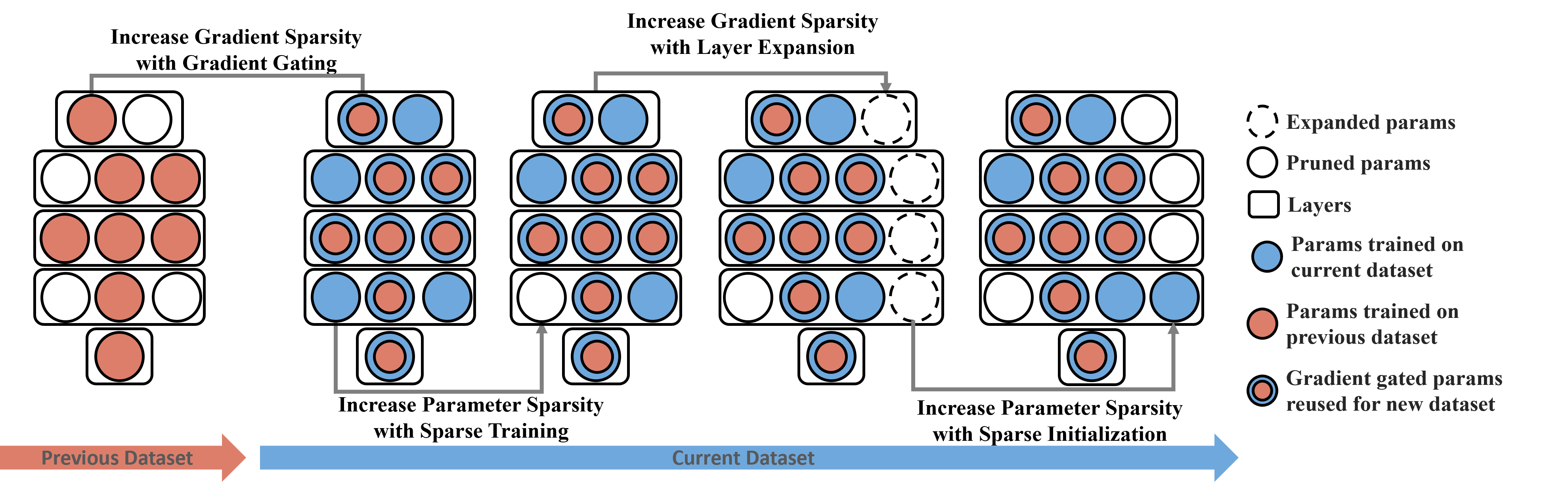}}
  \\
  \caption{SparseGrow training process overview. Due to the high complexity or dissimilarity of the new dataset (blue), the model's capacity limits its performance. Hence, there is a need to expand the model to enhance its adaptability and to reserve space for future data. Sparse training with freezing is used throughout the training process. Sparsity level and freeze mask updates as model expands.
}
     \label{figmethod}
  \end{center}
\end{figure*}

\section{Related Works}
\label{sec:formatting}

\subsection{Growing Approaches}
There are three main strategies for model growth: 1) Layer Expansion - this involves widening individual layers to expand the model's width, potentially increasing gradient sparsity during subsequent training sessions. 2) Lateral Connection - this strategy introduces new lateral layers that are connected to neighboring layers. The values from the lateral layers are then added to the main pathway, which does not necessarily increase the sparsity of gradients for upcoming tasks. And 3) In-depth Growth - this approach enhances the model's depth by adding more hidden layers. In continual learning, these methods are primarily used in task-specific scenarios, although there are a few exceptions in task-agnostic settings. However, these strategies do not directly address the issue of growth-induced forgetting.

\subsubsection{Lateral Connection}

Lateral connection grows new modules or layers in parallel with existing layers and allows connections with previous layers.

For instance, Progressive Neural Nets (PNNs) \cite{rusu2016progressive} statically grow the architecture with randomly initialized modules while retaining lateral connections to previously frozen modules. However, this method is limited to specific simple networks.
Schwarz et al. \cite{schwarz2018progress} use randomly initialized active lateral columns (1x1 conv) to learn new tasks by connecting them to lateral columns that store previous knowledge. They also distill knowledge from the active column to knowledge base layers. This method is applicable to Conv2d layers but is limited to specifically designed simple networks and lacks generality.
Zhang et al. \cite{zhang2020regularize} adopt AutoML-based model growing with both lateral connections and in-depth growth. They also use knowledge distillation to compress the model after learning a task. This method is applied to complex networks like VGG. However, using AutoML or similar methods often requires defining the layers for growth during the initial model definition, limiting its applicability. In-depth and lateral growth can also lead to more forgetting.
Ardywibowo et al. \cite{ardywibowo2022varigrow} propose VariGrow, which detects if a new task is arriving through an energy-based novelty score. If the novelty score is high and the sample is "detected" as a new task, VariGrow grows a new expert module to handle it. Otherwise, the sample is assigned to one of the existing experts that is most familiar with it. Unlike other methods, VariGrow is task-agnostic continual learning and does not require task identification during inference, making it suitable for class and domain incremental continual learning. However, this method can prevent the update of old modules when encountering similar data.
In Hu et al. \cite{hu2023dense}, given a new task $t$, a new branch called the task $t$ expert is added while freezing existing experts. This is achieved by introducing dense connections between the intermediate layers of the task expert networks.
Li et al. \cite{li2019learn} propose a hybrid solution for lateral connection and layer expansion, involving three operations: 'new', 'adaptation', and 'reuse'. The 'new' operation introduces a randomly initialized 3x3 layer trained from scratch. In the 'reuse' strategy, existing frozen weights are reused. The 'adaptation' strategy adds a 1x1 convolution layer in parallel to a 3x3 convolution layer, keeping the original 3x3 kernel fixed while learning the parameters of the 1x1 adapter. However, this hybrid solution is limited to specially designed Conv2d networks and lacks generality.
These methods demonstrate the potential of lateral connections for CL but highlight the need for more general approaches with less growth-induced forgetting applicable for complex networks.

\subsubsection{Layer Expansion}

DEN \cite{yoon2017lifelong} uses layer expansion in a top-down manner, growing every layer if the loss does not meet a threshold.
Similar to DEN, Hung et al. \cite{hung2019compacting} expand the number of filters (weights) for new tasks. Moreover, they adopt gradual pruning to compact the model. However, different from our sparse growing method, their pruning is not data-driven, and their method requires manual help of task IDs during inference, making it unsuitable for domain incremental or class incremental learning.
Ostapenko et al. \cite{ostapenko2019learning} introduce Dynamic Generative Memory (DGM) and expand the same number of neurons used in a layer in the generator of a GAN for the scalability of rehearsal. The scope of rehearsal is expanded.
Geng et al. \cite{geng-etal-2021-continual} expand the hidden size $H_k$ for the $k$-th task from $H_{k-1}$ by the pruning ratio of task $k-1$. After random initialization, on-data finetuning is performed for the newly added parameters.
Yang et al. \cite{yang2021grown} grow a randomly initialized expanded filter and concatenate it to the network.
Xu et al. \cite{xu2018reinforced} adaptively expand each layer of the network when the $t$-th task arrives. This method is also applicable for simple convolutional networks and fully-connected networks.
Yan et al. \cite{yan2021dynamically} expand the model with new parameters by creating a separate feature extractor $F_t$ for incoming data and taking a uniform distribution as the prior distribution. This method is also applicable for class incremental learning.
Our work could further exploit layer expansion's potential by reducing growth-induced forgetting.

\subsubsection{In-Depth Growth}

Some non-continual learning methods, such as Wen et al. \cite{wen2020autogrow} and Yuan et al. \cite{yuan2021growing}, increase the depth of the neural network in the hidden layer to achieve faster convergence and efficiency.
Yan et al. \cite{yan2021dynamically} also increase the depth of the neural network in the hidden layer to achieve faster convergence and efficiency. They use random initialization, which can help escape from a bad starting point.
Yuan et al. \cite{yuan2021growing} propose a budget-aware growing process that starts from a small, simple seed architecture and dynamically grows and prunes both layers and filters to make the network wider and deeper. They also adopt the initialization of ResNet or VGG.

In the field of continual learning, a few papers also utilize in-depth growth to increase model capacity:
Kozal et al. \cite{kozal2023increasing} add new layers on top of existing ones.
Zhang et al. \cite{zhang2020regularize} adopt AutoML-based model growing in both width and depth by adding lateral layers and in-depth layers. They also use knowledge distillation to compress the model after learning a task.
Nevertheless, adding new layers in depth within the neural network can have a profound impact on the model's structure compared to growing in width, leading to more severe multi-model forgetting and altering the learned features of previously acquired tasks. This effect undermines the network's ability to retain prior acquired knowledge and, therefore, is not suitable for continual learning.

\subsection{Overcoming Growth-Induced Forgetting}
The issue of growth-induced forgetting remains inadequately tackled, with only a limited number of studies focusing on this specific task-agnostic continual learning scenario. Notably, in task-agnostic CL, \cite{madaan2023heterogeneous} introduce Quick Deep Inversion to recover previous task visual features and improve distillation. Additionally, \cite{ostapenko2019learning} extend the neuron count in GAN generators to facilitate scalability through rehearsal. \cite{yan2021dynamically} freeze prior learned representations and introduce new feature dimensions from a trainable feature extractor. VariGrow \cite{ardywibowo2022varigrow} identifies new tasks and develops specific expert modules to handle them. Our work aims to confront the growth-induced forgetting issue at its root level, proposing novel strategies to enhance knowledge retention and address the growth-induced forgetting issue comprehensively.
 Overcoming growth-induced forgetting by handling its root causes and improving growth methods remains unexplored.

\section{Problem Formulation}
This study aims to enhance model adaptation to new data by expanding model capacity while addressing growth-induced forgetting in task-agnostic scenarios, covering domain-incremental and class-incremental learning.
Given a sequence of non-iid datasets $\{ D_1, D_2, \ldots, D_t \}$ and a model with weight $W$ trained on previous datasets, where $W_t$ are the model weights at time $t$.
Accessing only the current dataset $D_t$ at time $t$, the goal is to grow the model capacity to improve accuracy on $D_t$ and future datasets while minimizing performance degradation on previous datasets. This is formalized as:

\begin{equation}
\text{minimize } J(W_t) =\sum_{i=1}^{t-1} L(D_i; W_i) +  L(D_t; W_t) 
\end{equation}

where $J(W_t)$ is the objective function at time $t$, $\sum_{i=1}^{t-1} L(D_i; W_i)$ is the cumulative loss on previous datasets, and $L(D_t; W_t)$ is the loss on the current dataset.

\section{Methodology}

\subsection{Increased Gradient Sparsity}
Increasing the sparsity of model training gradients is an important way to reduce model forgetting\cite{mirzadeh2022wide}. When training on new tasks, updating only a small number of model parameters helps stabilize the network and reduces the forgetting of previously learned tasks. Common methods for regularization and structure can limit significant changes in model parameters, which reduces the variability of model gradients and thus minimizes forgetting. The techniques of layer expansion and gradient gating that we utilize also aim to prevent catastrophic forgetting by increasing the sparsity of model gradients.
As the network grows wider, the gradients become sparser, meaning fewer parameters need to be adjusted, which helps minimize forgetting of previously learned tasks.

\subsubsection{Layer Expansion}
Layer expansion is an effective growth method that minimizes forgetting while allowing for more general and detailed development without the need for manual selection of growth locations. Unlike lateral connections or extensive growth methods, this approach maintains the functionality of the model and is particularly well-suited for complex neural networks.

In this approach, we consider a neural network with a set of parameters denoted as \( W_l \) for \( 1 \leq l \leq C \), where \( W_l \) represents the parameter matrix at layer \( l \) and \( C \) is the total number of layers. Each layer in the neural network—such as Conv2D, MLP, or Batch Norm—undergoes an expansion technique. 

Given an original weight tensor \( W \in \mathbb{R}^{(c_o \times c_i)} \) and an original bias tensor \( b \in \mathbb{R}^{(c_o)} \), where \( c_o \) is the number of output channels and \( c_i \) is the number of input channels, these tensors are expanded into \( W_{exp} \) to accommodate the desired expansion.

\begin{equation}
W_{exp} \in \mathbb{R}^{(c_i + m, c_o + n)}; \quad b_{exp} \in \mathbb{R}^{(c_o + n)}.
\end{equation}

In this equation, \(m\) and \(n\) denote the expanded channel numbers. The expanded layer \(l\) now has \(c_i + m\) input channels and \(c_o + n\) output channels. Weights from the existing layer are transferred to their original positions within the new layer, while the expanded portion is initialized. The preservation and pruning masks are updated accordingly. After the expansion, the weights and biases of the expanded layer are updated as follows:

\begin{equation}
\begin{aligned}
W_{exp}[:c_o, :c_i] &= W \\
b_{exp}[:c_o] &= b
\end{aligned}
\end{equation}

Expansion across the entire model follows a systematic rule ensuring the alignment of input and output channels across consecutive layers. This expansion logic, consistent with structures like ResNet blocks and skip connections, maintains information flow by expanding involved layers while upholding the correspondence between input and output channels. Each layer ($l$ ranging from $1$ to $C$) is expanded by adding $n$ extra channels, aligning input and output sizes between adjacent layers:

\begin{equation}
c_o^{(l)} \rightarrow c_o^{(l)} + n; \quad 
c_i^{(l+1)} \rightarrow c_i^{(l+1)} + n 
\end{equation}

This gradual growth in capacity allows the network to capture more complex patterns and representations as the number of channels expands, while maintaining the necessary consistency for effective information propagation.

\subsubsection{Gradient Gating}
During training, gradients only influence non-preserved parameters, while preserved parameters remain fixed. 

The gradient gating mechanism can be described by the following equation:

\begin{equation} 
\nabla W_{\text{gated}} = (1 - P) \odot \nabla W
\end{equation}

In this equation, $\nabla W_{\text{gated}}$ represents the gated gradients, while $\nabla W$ contains the original gradients. The variable $P$ is the preservation mask, where a value of 1 indicates that the gradient is preserved, and a value of 0 indicates that it is trainable.

This approach is designed to preserve specific knowledge within the model, which is particularly useful in scenarios that require task-agnostic adaptation while maintaining previously learned information.

After training on each dataset, a preservation mask \( P \) is generated by aggregating the important learned parameters. 

Since unimportant parameters are pruned during sparse training and initialization, the preservation mask is computed as follows:

\[
P_{ij} = I\left(\left|W_{ij}\right| > 0\right),\quad \forall i\in[1,c_o],\forall j\in[1,c_i]
\]

In this equation, \( I \) is the indicator function, \( c_o \) is the number of output channels, \( c_i \) is the number of input channels, and \( W_{ij} \) represents the \( j \)th weight associated with the \( i \)th output neuron. 

To prevent gradients from affecting preserved parameters in expanded layers, we implement gradient gating as follows:

\[
W_{exp}^* = \underset{W}{\arg\min} L(D;W_{exp}) \odot (1 - P)
\]

In this equation, \( L(D;W) \) denotes the loss function applied to the current dataset, and \( W_{exp}^* \) is the optimal value of the expanded weights. The \( \odot \) operation performs element-wise gradient gating, where zero entries in \( (1-P) \) ensure that updates do not occur for the preserved weights.
Additionally, the preservation mask will also be expanded.

\subsection{Increased Parameter Sparsity}

Introducing increased parameter sparsity in continual learning aims to better control model plasticity, improve generalization, and enhance model adaptability. By leveraging data-driven pruning techniques, these methods provide a way to enhance parameter sparsity.

Sparse Training and Sparse Initialization are innovative strategies designed to increase parameter sparsity as tasks evolve. By utilizing data-driven pruning, these methods enable the network to dynamically adjust its parameter sparsity based on the specific characteristics of the data being processed. This approach not only promotes a more efficient allocation of model resources but also improves the network's resilience to catastrophic forgetting.

\subsubsection{Sparse Training}

The sparsification mechanism operates inherently within each layer, automatically regulating the levels of sparsity during training while maintaining the model's performance. 

Sparsification is achieved through a sparse mask $M_{ij}$ determined by trainable thresholds $t \in R^{c_o}$:

\begin{equation} \label{eq:sparse_gate}
M_{ij} = I\left(\left|W_{ij}\right| > t_i\right),\quad \forall i\in[1,c_o],\forall j\in[1,c_i]
\end{equation}

The threshold mechanism operates on a neuron basis for fully-connected layers (using threshold \( t_i \) for each output neuron) and on a filter basis for convolutional layers (using threshold \( t_i \) for each output channel).

Thresholding enables fine-grained sparsification while maintaining structural integrity. During forward passes, sparse operations are performed using the following matrix multiplication: \( W \odot M \). The threshold parameters \( t \) are dynamically optimized during training through a dual-phase process. To promote high sparsity in neural networks, a regularization term is introduced as follows:

\begin{equation}\label{eq3}
\mathcal{R} = \sum_{i=1}^{c_o} e(-t_i), \quad L_s = \sum_{i=1}^{C} \mathcal{R}_i
\end{equation}

In this context, \( e(-x) \) acts as a threshold regularizer that penalizes the threshold values \( t \) from becoming excessively large or too small. The training loss function integrates \( L_s \) to enable the training of a sparse neural network directly using backpropagation.

The complete optimization objective combines the task loss with the sparsity regularization:

\begin{equation}\label{eq:joint_optim}
W^*,t^* = \underset{W,t}{\arg\min} \left[L(D;W) + \alpha L_s\right] \odot (1 - P)
\end{equation}

Here, $\alpha$ controls the trade-off between sparsity and training accuracy. The exponential regularization \( e(-x) \) progressively penalizes low thresholds, prevents threshold explosion through asymptotic decay, and allows for layer-wise adaptation using independent \( t \) vectors.

When training on sequential datasets, the sparsification process is reinitialized for each new task. This approach allows new sparsity patterns to be formed for novel data while retaining the parameter configurations learned from previous tasks through the preservation mask \( P \). Importantly, the threshold parameters will also be expanded during the process of layer expansion.

\subsubsection{Sparse Initialization}

Sparse initialization involves employing random initialization with a sparse on-dataset warm-up phase.
While random initialization of grown parameters enables better adaptation to new data, it will cause higher growth-induced forgetting. We further apply on-dataset sparse initialization to fill the gap of randomly initialized grown parameters to the learned data pattern, therefore ensuring less forgetting.

Proper initialization of new parameters is crucial for adaptability and less forgetting, with zero-init hindering its update and random-init causing heavy growth-induced forgetting. To address this, we propose a simple yet effective approach: random and then on-data initialize the expanded part for several epochs to adapt it to the current distribution at the final phase of training.
The random initialization for a the expanded weight parameter $W_{exp}[c_o, :]$, the corresponding bias $b_{exp}$ and previous mentioned threshold vector $t_{exp}$ in a neural network layer is defined as:
\begin{equation}
\begin{aligned}
W_{exp}[c_o, :]_{\text{init}} &\sim \mathcal{N}\left(0, \frac{2}{n_{\text{in}}}\right) \\
t_{exp}[c_o, :]_{\text{init}}, \quad b_{exp}[c_o, :]_{\text{init}} &\sim \mathcal{N}\left(0, \frac{2}{n_{\text{in}}}\right)
\end{aligned}
\end{equation}

where $\mathcal{N}$ denotes the normal distribution, and $n_{\text{in}}$ represents the number of input units.
We then selectively warm-up the randomly initialized layer with the current dataset using sparse training to adapt it to the current distribution.
Sparse initialization involves updating the expanded parameters $W_{exp}$ and $b_{exp}$ using the current dataset by minimizing the sparse expanded loss function $L\left(D;W\right)+\alpha\ L_s$.
During this, we update the expanded parameters by computing their gradients with respect to the sparse expansion loss and freezing technique, thus applying an optimization algorithm:
\begin{equation}\label{eq5}
W_{exp}^\ast,t_{exp}^\ast=\underset{W,t}{\arg\min}[L\left(D;W_{exp}\right)+\alpha\ L_s] \odot (1 - P)
\end{equation}

This process allows the expanded part to adapt to the learned distribution, preventing the overwriting of previously learned knowledge while incorporating the expanded parameters' capacity for better adaptation and faster learning.

\begin{algorithm}[t]
\caption{Increasing Gradient Sparsity}
\begin{algorithmic}[1]
\REQUIRE Parameter matrix $W$, threshold vector $t$, sequence of datasets $\{D_0, D_1, \ldots\}$, expansion size $n$, sparsity hyperparameter $\alpha$
\FOR{dataset $D_k \in \{D_0, D_1, \ldots\}$}
  \FOR{epoch}
    \IF{final epochs \AND expansion phase}
      \FOR{layer in model}
        \STATE Create expanded parameter matrix $W_{exp} \in R^{(c_o + n, c_i + n)}$
        \STATE  Initialize them and threshold $W_{exp}[c_o, :]\text{init}$
      \STATE Copy old weight $W_{exp}[:c_o, :c_i] = W$
      \ENDFOR
    \ENDIF
    \FOR{training step $t$}
      \STATE Compute loss: $L =L(D;W_{exp}) + \alpha L_r$
      \IF{$k == 0$} \STATE Gradient decent with sparse parameter: $W_{exp}^*,t_{exp}^* = \arg\min [L(D;W_{exp}) + \alpha L_s]$
      \ELSIF{$k > 0$} \STATE Sparse gradient decent with sparse parameters: $W_{exp}^*,t_{exp}^* = \underset{W,t}{\arg\min} \left[L(D;W_{exp}) + \alpha L_s\right] \odot (1 - P)$
      \ENDIF
    \ENDFOR
  \ENDFOR
  \STATE Update preservation mask: $P = I\left(\left|W_{exp}\right| > 0\right)$
\ENDFOR
\end{algorithmic}
\label{alg:sparse_adapt}
\end{algorithm}

\begin{algorithm}[t]
\caption{Increasing Parameter Sparsity}
\begin{algorithmic}[1]
\REQUIRE Parameter matrix $W$, sequence of datasets $\{D_0, D_1, \ldots\}$, threshold vector $t$, sparsity hyperparameter $\alpha$
\FOR{dataset $D_k \in \{D_0, D_1, \ldots\}$}
  \FOR{layer $\ell \in \text{model}$}
    \STATE Initialize threshold $t \gets t_0$
  \ENDFOR
  \FOR{epoch}
    \FOR{training step $t$}
    \FOR{layer in model}
      \STATE Update sparsity masks: $M = I(|W| > t)$ 
      \STATE Update sparsified weights: $W = W\odot M$
      \ENDFOR
      \STATE Compute loss: $L = L(D;W) + \alpha L_s$
    \ENDFOR
  \ENDFOR
\ENDFOR
\end{algorithmic}
\label{alg:sparse_adapt}
\end{algorithm}

\section{Experiments and Results}

\subsection{Datasets}
We evaluate on diverse datasets: Permuted MNIST, FreshStale, and DomainNet for domain-incremental tasks; CIFAR-10 and class-incremental MNIST for class-incremental evaluations.
\begin{itemize}
    \item \textbf{Permuted MNIST} Variably permuted MNIST images with no shared features across permutations, each set containing 70,000 images.

    \item \textbf{Class-incremental MNIST} Divides MNIST into distinct class groups with varying sizes and overlaps.
    \item \textbf{CIFAR-10} 60,000 color images in 10 classes grouped unevenly for class-incremental experiments.
    \item \textbf{FreshStale} \cite{joseph2021food} 14,683 images of fruits and vegetables labeled fresh or stale, totaling around 2GB.
    \item \textbf{DomainNet} \cite{peng2019moment} Dataset with different domains such as real photos, clipart, quickdraw, and sketch, varying data sizes (48K - 172K images) and 345 classes per domain.
    
\end{itemize}

\subsection{Evaluation Metrics}
For a principled evaluation, we adopt the following evaluation metrics \cite{lopez2017gradient}:

\begin{itemize}
\item Average Accuracy:
$AAC=\frac{1}{T}\sum_{i=1}^{T}R_{T,i}$

\item Backward Transfer:
$BWT=\frac{1}{T-1}\sum_{i=1}^{T-1}\left(R_{T,i}-R_{i,i}\right)$

\item Forward Transfer:
$FWT=\frac{1}{T-1}\sum_{i=2}^{T-1}{R_{i-1,i}-\bar{b_i}}$
\end{itemize}

where $R_{T,i}$ represents the test accuracy of each dataset after all datasets are learned.
$R_{i,j}$ is the test accuracy on dataset $t_j$ after observing the last sample from dataset $t_i$.  $\bar{b}$ is test accuracy at random initialization. 
The primary evaluation criterion is the average accuracy (AAC), higher the better. When AAC is the same, larger BWT or FWT is superior. We assess parameter efficiency through parameter calculations (Params).

\subsection{Experiment Settings}
All baselines are applied on ResNet-18 as base network. To guarantee reproducible results, we set the seed value as 5 for the random function of Numpy, python Random, Pytorch, Pytorch Cuda, and set Pytorch backends Cudnn benchmark as False, with Deterministic as True, so that multiple calls to those operations will produce the same result, given the same inputs.
For fair comparison of different model growth technique, we increase Conv2d and MLP layers of base network uniformly. We random initialize grown parameters for baseline methods.
We expanded both MLP and Conv2d layers by increasing channels to $n=2$. For experiments overcoming growth-induced forgetting, the model expand once before the last task, where growth-induced forgetting is strongest. To demonstrate model scalability in Figure \ref{fig12mnist}, the models expand after the second task to highlight model growth's impact. We select hyperparameter $\alpha$ based on the volume of data and training epochs. A suitable choice for $\alpha$ could be the reciprocal of the total training iterations (data $\times$ epochs).

\subsection{Baselines}
Existing works on model growth such as DEN and PNN use specially designed structures, mainly for task-specific settings. To ensure a fair comparison, we abstract their model growth methods and apply them to the same ResNet-18 network.
\textbf{(1) Classic and recent baselines without model growth:}

    \begin{itemize}
        \item \textbf{SGD} \cite{bottou1991stochastic}: Basic model trained with stochastic gradient descent.
        \item \textbf{EWC} \cite{kirkpatrick2017overcoming}: Elastic Weight Consolidation, a regularization technique.
        \item \textbf{LwF} \cite{li2017learning}: Knowledge distillation approach for retaining past knowledge.
        \item \textbf{PRE-DFKD} \cite{binici2022robust}: Recent data-free rehearsal strategy using VAE-based knowledge distillation.
        \item \textbf{PackNet} \cite{mallya2018packnet}: Neuron pruning based on known task numbers.
        \item \textbf{AdaptCL} \cite{zhao2023adaptcl}: Recent task-agnostic method using dynamic pruning and freezing.
    \end{itemize}
    
  \textbf{(2) Baselines for model growth comparison:}
    \begin{itemize}
        \item \textbf{SGD+LayExp:} Layer Expansion on a base network trained with SGD.
        \item \textbf{SGD+IDGrow:} In-Depth Growth applied to base networks with SGD.
        \item \textbf{SGD+LatConn:} Lateral Connections used on base networks with SGD.
        \item \textbf{SGD+LayExp+ODInit:} Layer Expansion and On-Dataset Initialization incorporated with SGD.
    \end{itemize}
    
\textbf{(3) Baselines with layer expansion and on-data initialization:}
    \begin{itemize}
        \item \textbf{EWC+LayExp:} EWC with layer expansion, expanding the Fisher Information Matrix accordingly.
        \item \textbf{LwF+LayExp:} LwF merged with layer expansion, distilling knowledge post-expansion.
        \item \textbf{PRE-DFKD+LayExp:} Layer expansion applied alongside PRE-DFKD to base networks.
    \end{itemize}
\subsection{Comparison of Model Growth Techniques}

\begin{table}[!ht]
\small
\centering
\caption{Performance evaluation of different model growth methods in terms of average accuracy, backward knowledge transfer, forward knowledge transfer, and number of used parameters (Params $\times10^7$) after observing four domains of Permuted MNIST. }
\begin{tabular}{lcccc}
\toprule
Method & AAC↑ & BWT↑ & FWT↑ & Params↓ \\
\midrule
No Growth & 0.666 & -0.424 & 0.014 & 1.117 \\
Lateral Connection & 0.424 & -0.747 & 0.050 & 2.216 \\
In-Depth Growth & 0.373 & -0.814 & 0.133 & 1.993 \\
\textbf{Layer Expansion}& \textbf{0.679} & \textbf{-0.406} & 0.014 & 1.131 \\
\bottomrule
\end{tabular}
\label{tb1}
\end{table}

To evaluate the impact of different model growth strategies on performance metrics such as average accuracy, knowledge transfer (including catastrophic forgetting and growth-induced forgetting issues), and parameter efficiency, we compare the effects of layer expansion, lateral connections, in-depth growth, and no growth on a domain incremental permuted MNIST dataset containing four domains. 

The summarized results are presented in Table \ref{tb1}.
The findings show that, in comparison to no growth, layer expansion not only mitigates growth-induced forgetting but also promotes backward knowledge transfer, consequently increasing the overall model accuracy with only a 1.2\% rise in parameters. Conversely, other growth methods exhibit significant growth-induced forgetting, leading to reduced average accuracy. Furthermore, as both lateral connections and in-depth growth necessitate full-layer expansion for uniform growth across MLP and Conv2d layers, the model's parameter count nearly doubles.

\subsection{Comprehensive Comparison with More Tasks}

\begin{figure}[!ht]
  \begin{center}
{\includegraphics[width=\linewidth]{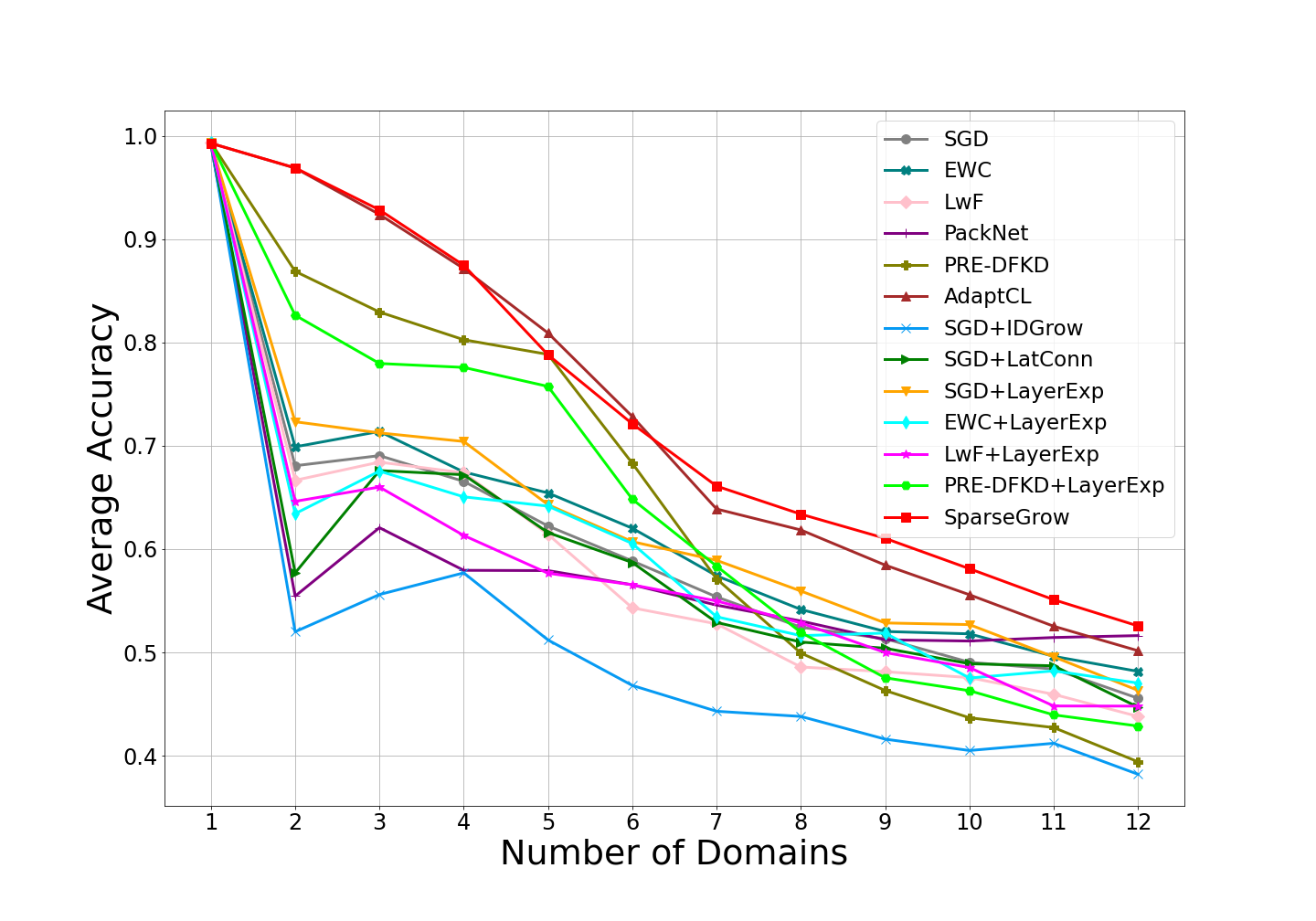}}
  \\
  \caption{Average accuracy fluctuation of different methods across an increasing number of observed domains. Methods include continual learning baselines, model growth techniques, and continual learning methods+LayExp. Rehearsal methods like LwF and PRE-DFKD initially decline when combined with LayExp, indicating potential unsuitability for directly applying LayExp and the risk of increased growth-induced forgetting; later, LayExp positive effects outweigh the negative impact of growth-induced forgetting as domains increase. SparseGrow excels in knowledge retention, improving effectiveness as the number of domains rises.
}
\label{fig12mnist}
  \end{center}
\end{figure}

\begin{table}[!ht]
\small
\centering
\caption{Comprehensive evaluation of average accuracy (AAC), backward transfer (BWT), forward transfer (FWT), and used parameters (Param $\times 10^8$) on twelve domains of Permuted MNIST. 
Comparison includes: 1) Simple baselines, 2) SGD with model growth techniques IDGrow, LatConn, LayExp and 3) Baselines with LayExp and ODInit 
}
    \begin{tabular}{l|cccc}
        \toprule
      Method &AAC↑& BWT↑&FWT↑&Param↓\\
        \midrule
    SGD                     & 0.4556  & -0.5739 & 0.0493 &1.1173 \\ 
    EWC                     & 0.4816  & -0.5454 & 0.0638 &1.1173 \\ 
    LwF                     & 0.3976  & -0.6387 & 0.0540 &1.1173 \\ 
    PackNet                 & 0.5163  & -0.5049 & 0.0659 &1.1173 \\ 
    PRE-DFKD                & 0.3941  & -0.6392 & 0.0774 &1.1173 \\ 
    AdaptCL        & 0.5016  & -0.2532 & 0.0865 &1.1173 \\ 
  \midrule
SGD+LatConn & 0.4474	& -0.6799	& 0.0779&2.2164\\

SGD+IDGrow	& 0.3822	& -0.6532	& 0.1418 &1.3662
\\
    SGD+LayExp                 & 0.4632  & -0.5658 & 0.0737 &1.1306\\
      SGD+LayExp+ODInit & 0.4683&-0.5598&0.0742&1.1306    
      \\ 
      \midrule
    EWC+LayExp                 & 0.4703  & -0.5571 & 0.0597 &1.1306 \\ 
    LwF+LayExp                 & 0.3941  & -0.6426 & 0.0680 &1.1306 \\ 
    PRE-DFKD+LayExp            & 0.4064  & -0.6248 & 0.0759 &1.1306 \\
EWC+LayExp+ODInit&0.4755&	-0.5498	&0.0594&1.1306\\
LwF+LayExp+ODInit&0.3971	&-0.6405&	0.0668&1.1306\\
P-DFKD+LExp+SInit&	0.4199	&-0.6196	&0.0747&1.1306\\    
  \textbf{SparseGrow(our)}              & \textbf{0.5256}  & \textbf{-0.2066} & \textbf{0.0974 }& \textbf{1.1054} \\ 
        \bottomrule
    \end{tabular}
    \label{tb12mnist}
\end{table}

\begin{table*}[!ht]
\small
\caption{Performance evaluation of continual learning methods in terms of average accuracy (AAC), backward knowledge transfer (BWT), and forward knowledge transfer (FWT) on the domain-incremental datasets of FreshStale and DomainNet. 
}
\centering
\begin{tabularx}{5.5in}{lXXXX|XXXX}
 \toprule
\multirow{2}*{Method} & \multicolumn{4}{c}{FreshStale}&\multicolumn{4}{c}{DomainNet}\\
\cmidrule{2-9}
&AAC↑ & BWT↑ & FWT↑ & Param↓& AAC↑ & BWT↑ & FWT↑ & Param↓ \\
\midrule
SGD+LayExp & 0.617 & -0.450 & 0.027 &1.131& 0.464 & -0.657 & 0.109 &1.131\\
EWC+LayExp & 0.694 & -0.360 & 0.039&1.131 & 0.463 & -0.661 & 0.109 &1.131\\
LwF+LayExp & 0.641 & -0.424 & 0.009 &1.131& 0.434 & -0.699 & 0.099&1.131 \\
PRE-DFKD+LayExp & 0.703 & -0.200 & 0.089 &1.131& 0.446 & -0.475 & 0.090 &1.131\\
\textbf{SparseGrow(our)} & \textbf{0.737} & -0.273 & 0.049 &\textbf{1.121}& \textbf{0.624 }& \textbf{-0.283} & 0.091&\textbf{1.120} \\
\bottomrule
\end{tabularx}
\label{tbfreshdomain}
\end{table*}


We conduct a comprehensive performance comparison of all baselines, including existing methods with and without model growth, on domain-incremental permuted MNIST datasets.
Figure \ref{fig12mnist} depicts average accuracy fluctuation with an increasing number of tasks or domains for various baseline methods. The comparison between SparseGrow and AdaptCL emphasizes the importance of appropriate model growth in enhancing adaptability facing more domains with only one expansion. Our model achieves the best performance, with its effectiveness improving as the domain number rises. 
Comparison among growth strategies suggests that layer expansion consistently improves model accuracy by increasing gradient sparsity, while lateral connections show unstable effects and in-depth growth mostly has negative impacts.
Notably, rehearsal methods like LwF and PRE-
DFKD exhibits an initial decline when combined with LayExp, suggesting such techniques may not integrate well with layer expansion directly, possibly leading to more growth-induced forgetting. Yet, the later positive effects of model capacity enhancement from expansion seem to surpass the negative impact of growth-induced forgetting as domains increase. LwF's average accuracy falls below SGD, suggesting that knowledge distillation methods may not be suitable for permuted MNIST that lack cross-domain similarity.
For PackNet, the model's average accuracy remains nearly constant as the number of domains increases. Incorporating on-data initialization into all baselines with layer expansion leads to improved AAC overall, indicating the effectiveness of this approach.

Table \ref{tb12mnist} compares the AAC, BWT, and FWT metrics of three categories of CL models after learning from 12 domains consecutively.
SparseGrow, in comparison to baselines, significantly mitigates overall forgetting issues, enhancing adaptability and surpassing SGD with LayExp in BWT by 64\% and AAC by 15\%, demonstrating better knowledge retention. On-data initialization, when combined with layer expansion, shows consistent improvements on layer-expanded baselines such as SGD, EWC, LwF and PRE-DFKD, in AAC, BWT, and FWT, showing it as a generic approach. PackNet, utilizing static pruning, exhibits advantages with a larger number of datasets; however, its reliance on known total task numbers makes it unfeasible in CL scenarios.
\begin{figure}[!ht]
  \begin{center}
  \
{\includegraphics[width=\linewidth]{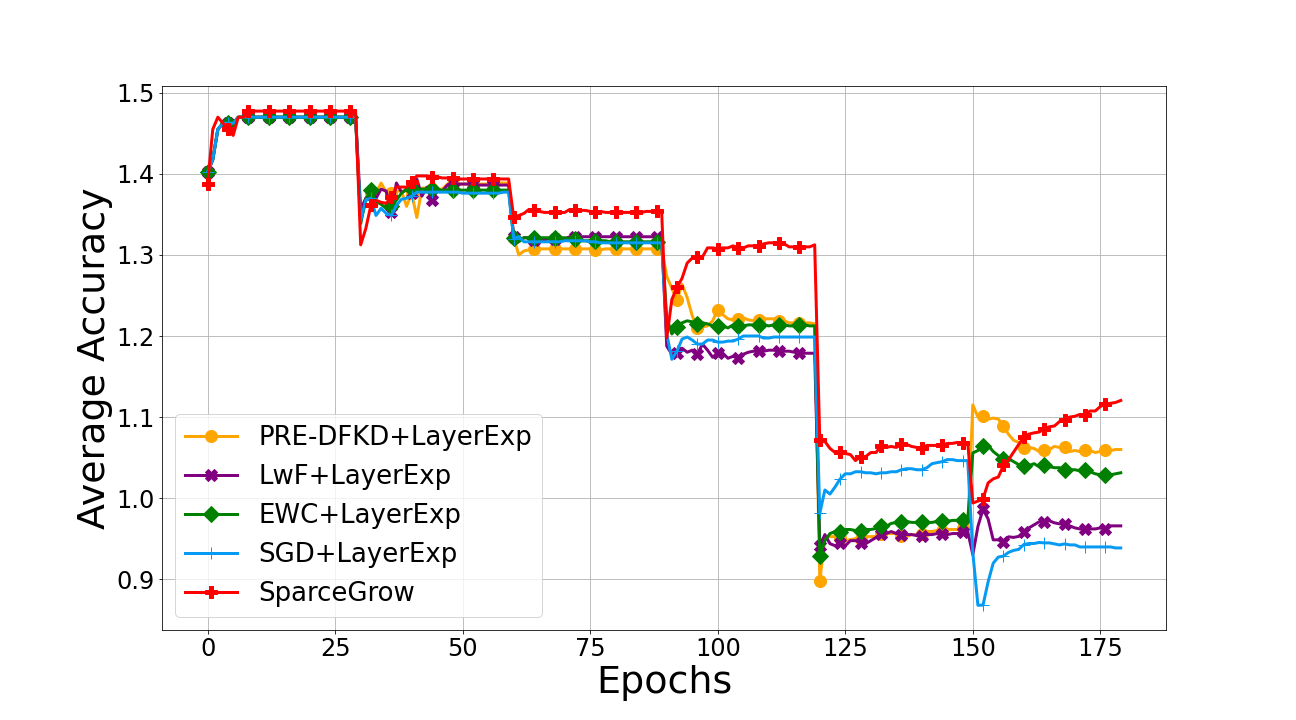}}
  \\
  \caption{Epoch-wise average accuracy of observed domains using different continual learning methods with layer expansion on FreshStale datasets with six sequential domains. 
}
     \label{figfresh}
  \end{center}
\end{figure}

\begin{figure}[!ht]
  \begin{center}
{\includegraphics[width=\linewidth]{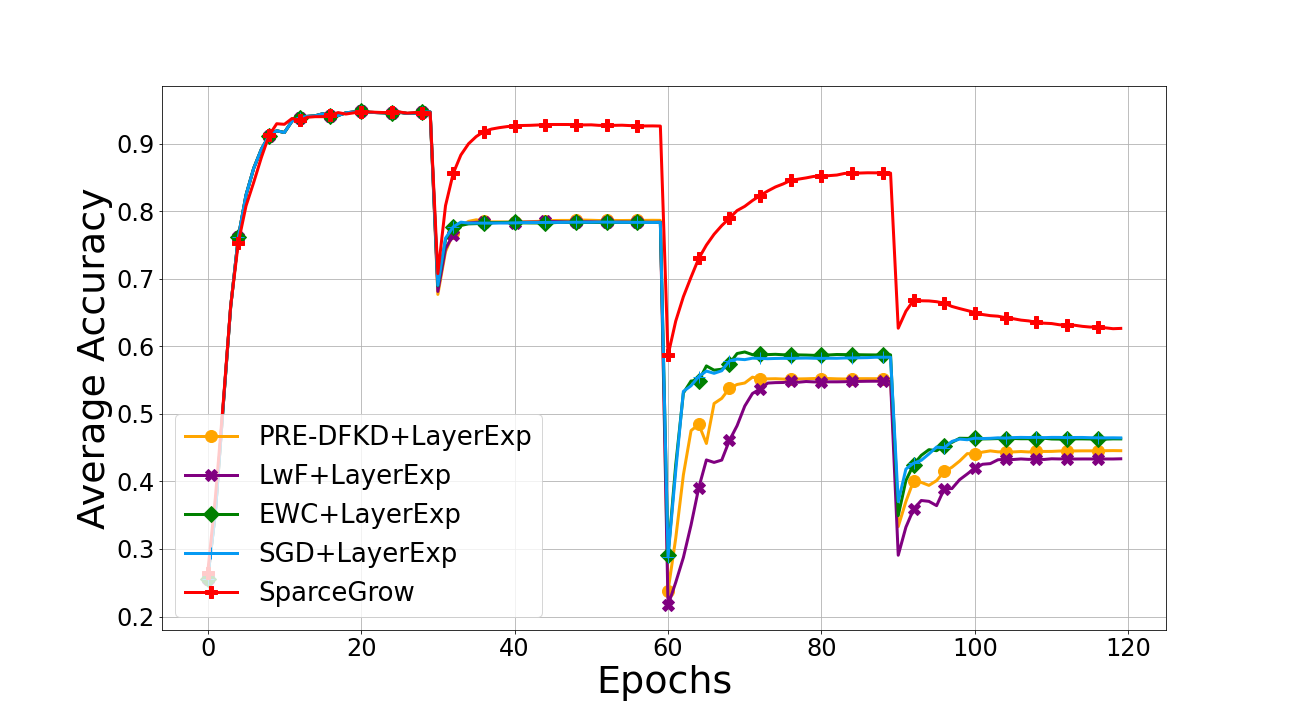}}
  \\
  \caption{Epoch-wise average accuracy of observed domains using different continual learning methods with layer expansion on DomainNet datasets with four sequential domains. 
}
     \label{figdomain}
  \end{center}
\end{figure}

\begin{table*}[!ht]
\scriptsize
\caption{Comparison of average accuracy (AAC), backward knowledge transfer (BWT), forward knowledge transfer (FWT), and each dataset's test accuracy using different continual learning methods in the class-incremental setting on MNIST and CIFAR-10 datasets.
}
    \centering
\begin{tabularx}{\linewidth}
{lXcXXXXXX|XccXXXXX}
        \toprule
\multirow{3}*{Method}&	\multicolumn{8}{c}{Class-Incremental MNIST}&
\multicolumn{8}{c}{Class-Incremental CIFAR-10}
\\
\cmidrule{2-17}
&\multirow{2}*{AAC↑}&
\multirow{2}*{BWT↑}&	\multirow{2}*{FWT↑}&
\multirow{2}*{Param↓}&
\multicolumn{4}{c}{Test Accuracy↑}&
\multirow{2}*{AAC↑}&
\multirow{2}*{BWT↑}&	\multirow{2}*{FWT↑}& \multirow{2}*{Param↓}&	\multicolumn{4}{c}{Test Accuracy↑}\\
\cmidrule{6-9} \cmidrule{14-17}
&	&	 &	 & & 	0,1,2&2,3,4,5&	0,3,6,7& 6,8,9
&	&	 &	 & & AP-D & AP,D-H&B,Do,S& S,T\\
\midrule

SGD+LayExp & 0.309 & -0.917 & 0.188 &1.131& 0.000 & 0.000 & 0.240 & 0.996
& 0.303 & -0.753 & -0.019 &1.131& 0.000 & 0.235 & 0.000 & 0.978\\
EWC+LayExp & 0.309 & -0.917 & 0.188 &1.131& 0.000 & 0.000 & 0.240 & 0.996
 & 0.113 & -0.179 & -0.095 &1.131& 0.200 & 0.250 & 0.000 & 0.000\\
LwF+LayExp & 0.309 & -0.917 & 0.187 &1.131& 0.000 & 0.000 & 0.239 & 0.997
& 0.305 & -0.752 & -0.023 &1.131& 0.000 & 0.239 & 0.000 & 0.980\\
PRE-DFKD+LayExp & 0.330 & -0.839 & 0.181 &1.131& 0.100 & 0.000 & 0.336 & 0.883 
& 0.303 & -0.691 & -0.040&1.131 & 0.000 & 0.237 & 0.000 & 0.976 \\
\textbf{SparseGrow(our)} & \textbf{0.412} & \textbf{-0.623} & 0.165&\textbf{1.114} & \textbf{0.279} & \textbf{0.081} & 0.307 & 0.981
& \textbf{0.322} & \textbf{-0.665} & \textbf{-0.018} & \textbf{1.101}&\textbf{0.042} & \textbf{0.311} & \textbf{0.079} & 0.855 \\

        \bottomrule
    \end{tabularx}%
    \label{tbclass}%
\end{table*}%

\subsection{Addressing Growth-Induced Forgetting Evaluation}

In our study assessing the efficacy of various continual learning strategies in tackling growth-induced forgetting across diverse applications, we focused on their layer-expanded versions, which inherently confront growth-induced forgetting challenges. The evaluations were conducted using the FreshStale and DomainNet datasets.
Figure \ref{figfresh} and Figure \ref{figdomain} visualize the epoch-wise average accuracy of different baseline methods on these datasets. Additionally, Table \ref{tbfreshdomain} presents the AAC, BWT, and FWT metrics for these applications.

Figure \ref{figfresh} illustrates that SparseGrow consistently achieves the highest epoch-wise average accuracy across six domains, ultimately exhibiting the best post-training average accuracy. All baseline methods exhibit enhancements in average accuracy and backward transfer compared to SGD. This improvement suggests that rehearsal, regularization, and structure-based methods hold promise in mitigating growth-induced forgetting resulting from layer expansion. Notably, SparseGrow stands out for significantly reducing model forgetting, ultimately achieving the highest average accuracy among the methods tested.
In Figure \ref{figdomain}, SparseGrow demonstrates exceptional effectiveness in countering growth-induced forgetting, surpassing other methods in mitigating the effects induced by layer expansion, particularly when compared to SGD.

\begin{figure*}[!ht]
  \begin{center}
{\includegraphics[width=\linewidth]{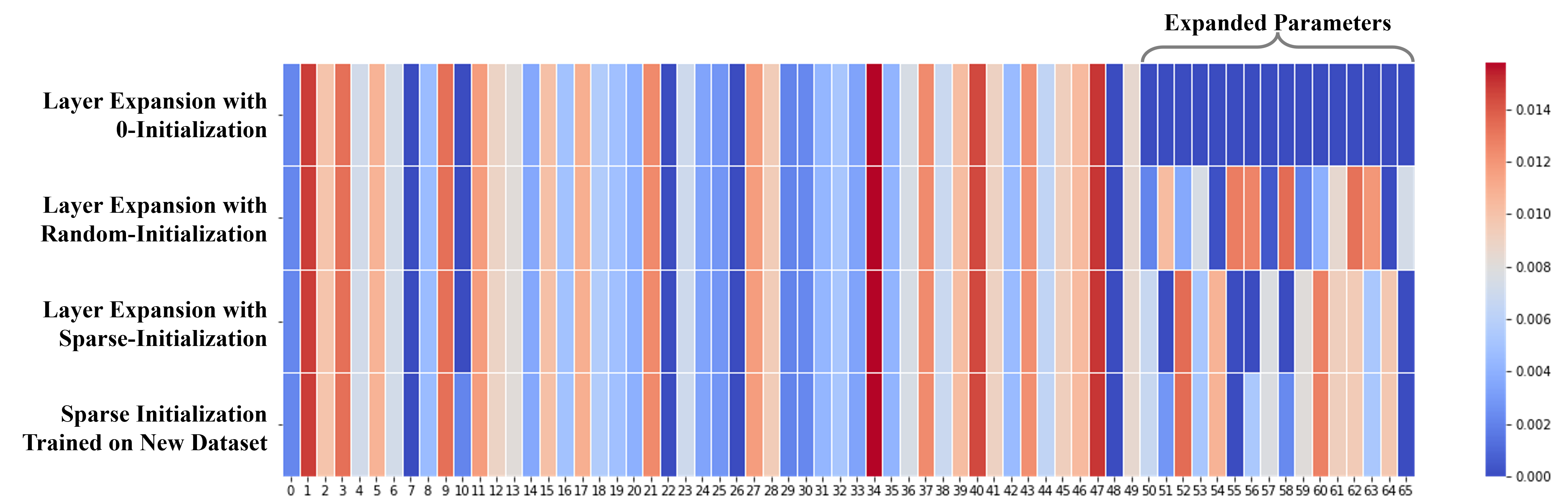}}
  \\
  \caption{Firing probability of the last 50 parameters within first Conv2d layer of ResNet18 on PermutedMNIST datasets. The row from top to bottom each represent 1) last 50 parameters before expansion, 2) after expansion of 2 channels with 0-initialized 16 parameters, 3) expansion with random initialization like existing works, 4) our on-data initialized sparse expansion with partial pruned values, and 5) our method trained on new dataset. 
}
     \label{figdomain}
  \end{center}
\end{figure*}

The results from Table \ref{tbfreshdomain} reveal that PRE-DFKD demonstrates the highest backward transfer, effectively alleviating forgetting in the FreshStale scenario. SparseGrow achieves the best average accuracy, with its BWT ranking second, indicating its optimal utilization of model expansion capacity to enhance adaptability to new data. Upon data analysis, we identified challenges in our model's performance on the Orange dataset post-training on Capsicum and Bitter Gourd datasets due to conflicting features.
For the DomainNet dataset, our model excels in overcoming forgetting and achieving the best average accuracy. The table underscores that existing regularization and rehearsal-based techniques show efficacy in addressing growth-induced forgetting, albeit with varying performance across datasets. In contrast, SparseGrow consistently mitigates growth-induced forgetting issues across these diverse applications.

\subsection{Class-Incremental Learning Setting Results}
In our exploration of task-agnostic continual learning scenarios, we carefully review the methods using class-incremental learning datasets, detailed in Table \ref{tbclass}. This experimental setup involves instances where similar inputs can lead to different classes assigned to various output layers, often resulting in rapid forgetting within a single epoch. Unlike conventional class-incremental research that trains baselines across multiple iterations, we opted to train each task for a minimum of 5 epochs, encompassing tens of thousands of iterations per epoch. Tasks were structured with varying numbers of classes to evaluate the methods' resilience.

Given the intensified forgetting challenges and dataset scale, techniques such as rehearsal and regularization methods encountered difficulties in maintaining accuracy effectively, exhibiting only marginal superiority over SGD. PRE-DFKD demonstrated improvement in overcoming forgetting compared to SGD, preserving more accuracy on previous tasks. However, it displayed weaker adaptability in the final task of class-incremental MNIST, possibly due to the increased information for rehearsal, leading to a struggle in balancing old and new knowledge.
SparseGrow emerged as a standout performer in terms of average accuracy and backward transfer on both datasets, underscoring its potential in mitigating forgetting in task-agnostic environments. While for CIFAR-10, SparseGrow also displayed reduced adaptability in the final task, potentially due to insufficient model capacity expansion (expansion once with two channels) in the class-incremental scenario that demands more model capacity. Notably, EWC encountered breakdowns during the last stages of training on CIFAR-10, possibly due to strong distribution shifts.

\subsection{Ablation Studies}

\begin{figure}[ht]
  \begin{center}
\includegraphics[width=3in]{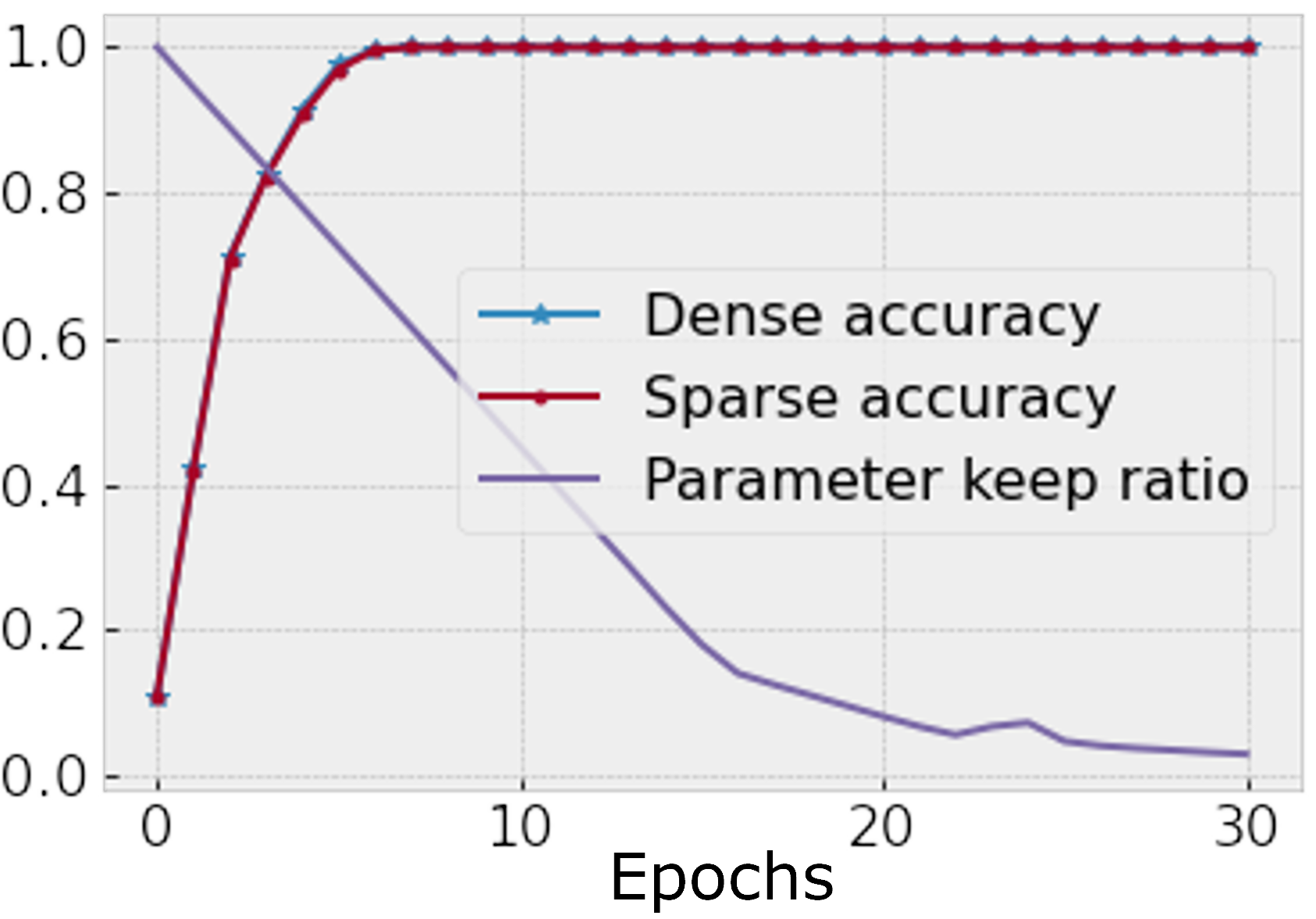}
\includegraphics[width=3in]{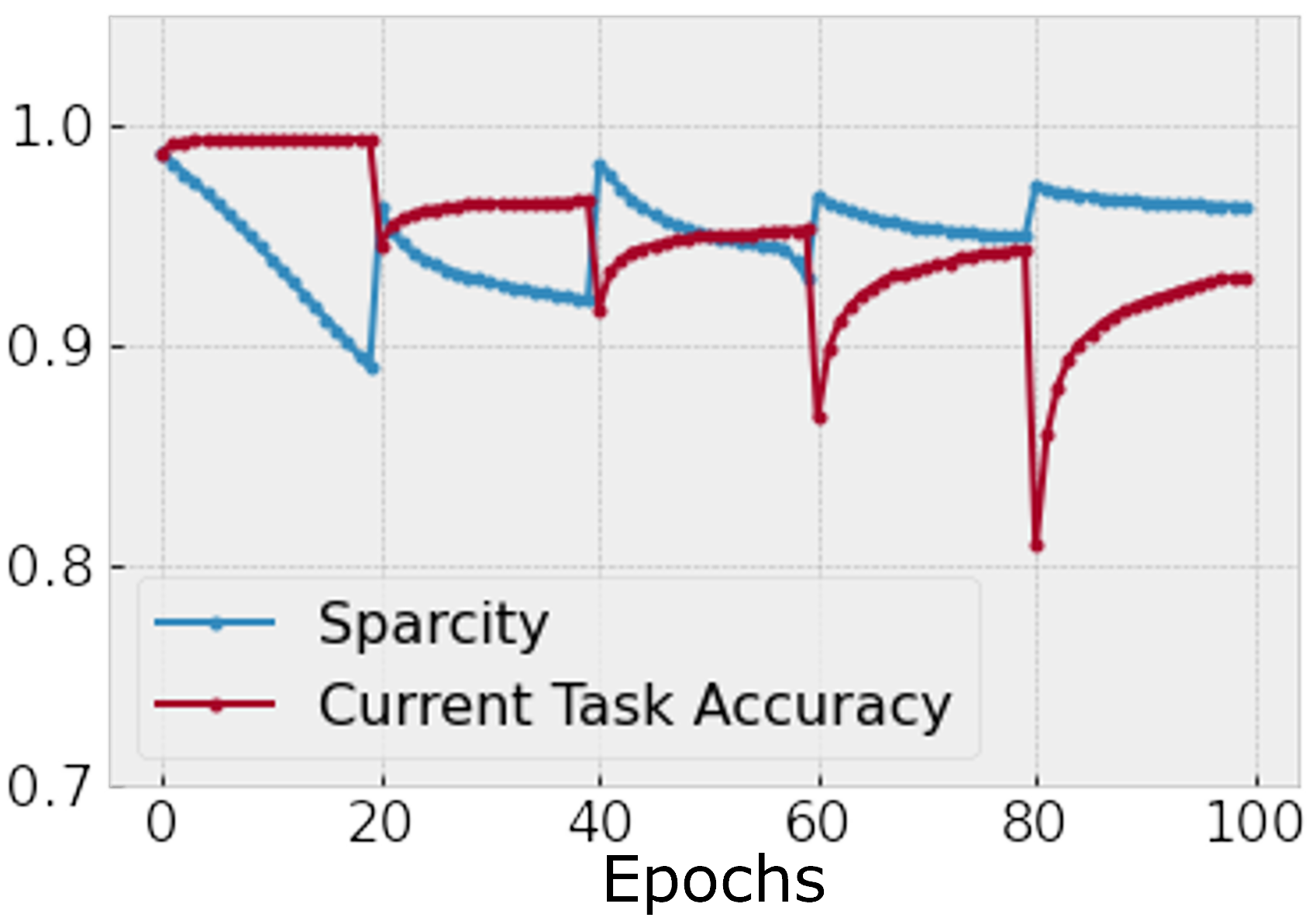}
  \caption{(a) Model on ResNet-18 remaining ratio and sparse accuracy compared with dense accuracy, using $\alpha=10^{-4}$ (b) Change of model parameter remaining ratio and current task accuracy of SparseGrow running on 5 tasks. (Best viewed in color)}
\label{figure10}
  \end{center}
\end{figure}

\begin{figure}[ht]
  \begin{center}
{%
\includegraphics[width=\linewidth]{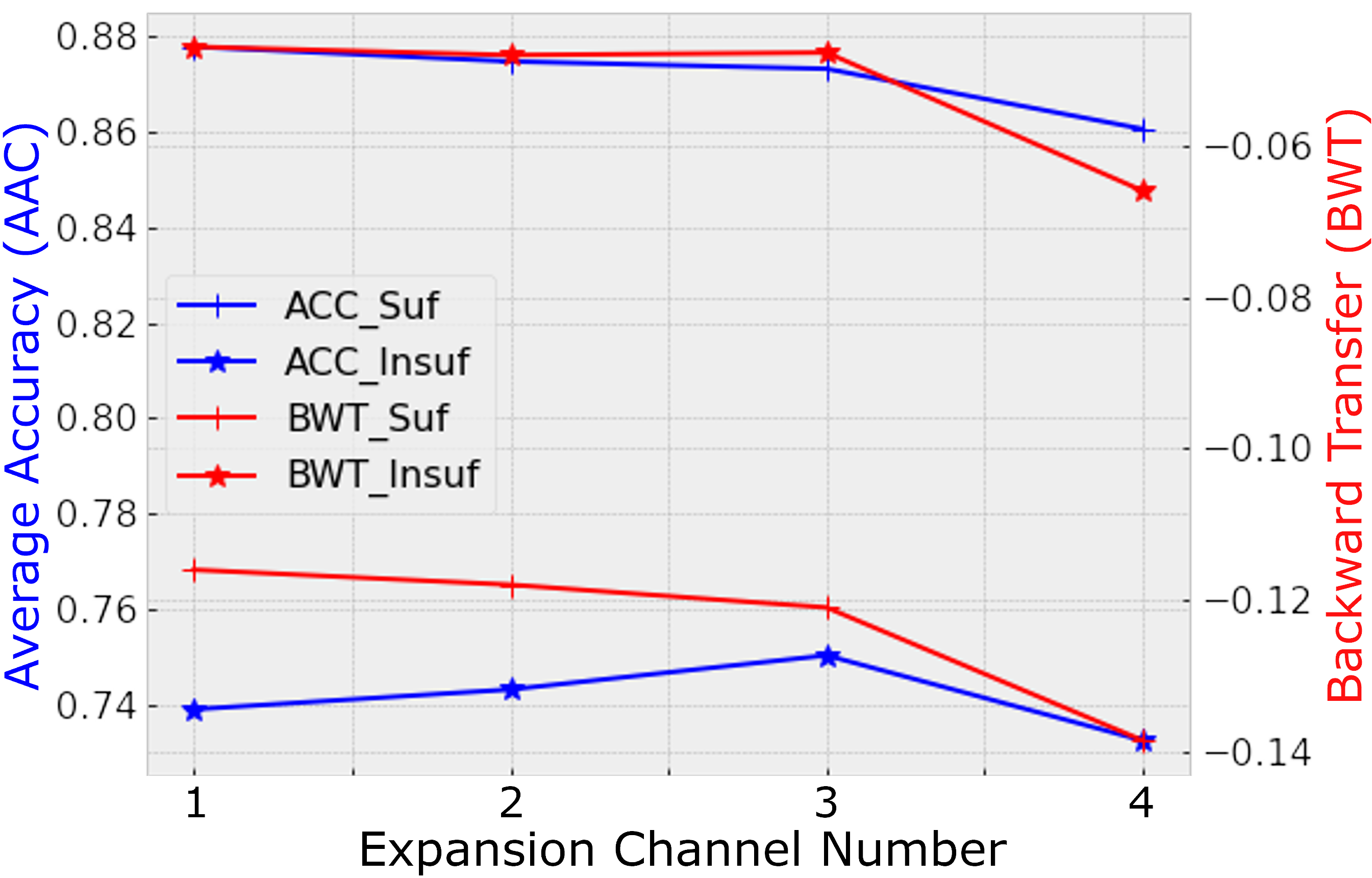}}
{%
\includegraphics[width=\linewidth]{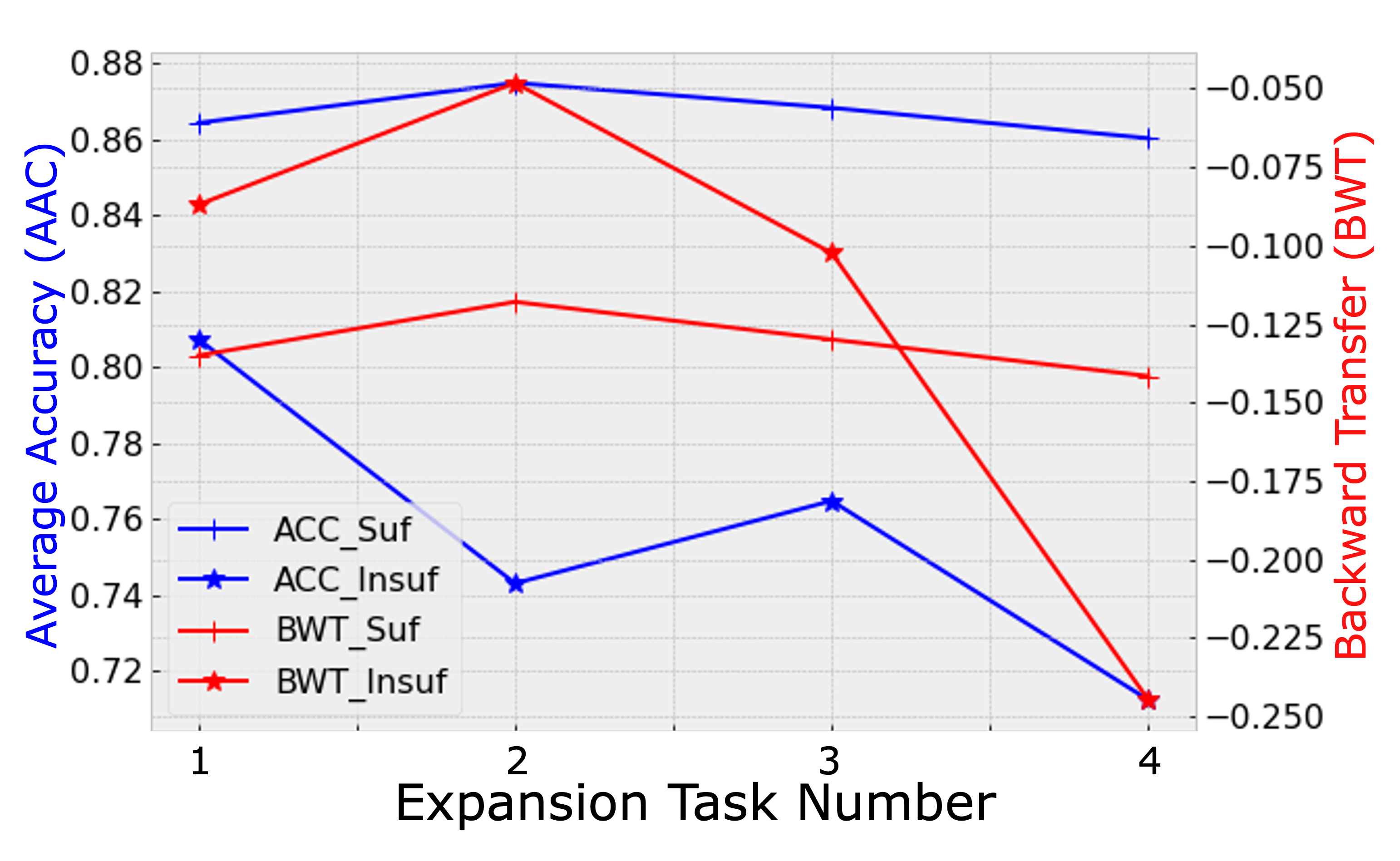}}
  \caption{Average accuracy and backward transfer of models with sufficient capacity (Suf) and insufficient capacity (Insuf) according to changes in (a) number of expansion channels $n$ and (b) expansion at which number of task.}
\label{figure11}
  \end{center}
\end{figure}

\textbf{Accuracy Preservation in Sparse Training:}

We investigate the impact of sparse training on training accuracy by analyzing the parameter remaining ratio alongside epoch-wise evaluation accuracy. In Figure \ref{figure10}(a), we compare a standard dense network with no sparse training to a network trained using neural sparse techniques. The results demonstrate that as parameters are progressively pruned, the sparse network achieves comparable accuracy to the dense network while utilizing less than 10\% of the parameters. This observation highlights the effectiveness of sparse neural training in maximizing model capacity without compromising training accuracy. Additionally, as depicted in Figure \ref{figure10}(b), SparseGrow dynamically adjusts sparsity levels based on training accuracy, showcasing a correlation between decreased model capacity and reduced accuracy on new tasks.

\textbf{Relationship Between Expansion Size, Time, and Model Capacity:}
To explore the influence of expansion hyperparameters on model performance concerning capacity sufficiency, we analyze average accuracy and backward transfer variations across different expansion sizes and times. Figure \ref{figure11}(a) indicates that an expansion involving 3 channels provides the most significant enhancement in model adaptability to new data, leading to decreased backward transfer and increased average accuracy for models with limited capacity. Conversely, Figure \ref{figure11}(b) reveals that expanding after task 2 results in minimal forgetting, as indicated by backward transfer metrics. For models with constrained capacity, expansion following task 1 yields the highest average accuracy, underscoring the potential of model expansion to augment capacity for fine-tuning or continual learning with pre-trained models. Notably, pre-trained models remain unaltered for regularization and do not store data for rehearsal, making them suitable candidates for capacity enhancement through expansion strategies.

\section{Conclusion}

Our study focuses on the crucial issue of growth-induced forgetting caused by improper model growth in task-agnostic continual learning. We identify layer expansion as a promising fundamental model growth technique, especially for pre-trained models. Our proposed SparseGrow approach is specifically designed to boost adaptability and knowledge retention of layer expansion. Experimental validations have demonstrated the effectiveness of our method in mitigating growth-induced forgetting and improving knowledge retention for incremental tasks. Future research could aim to optimize the timing of model growth and leverage techniques such as neural architecture search to further enhance the model's adaptability and overall performance with new data.







\bibliographystyle{elsarticle-harv} 
\bibliography{example}






\end{document}